\documentclass[12pt]{article}

\usepackage{amsfonts}
\usepackage{amsmath}
\usepackage{amssymb}

\usepackage{cite}

\usepackage{enumerate}
\usepackage{tikz}

\usepackage[colorlinks,linkcolor=blue,citecolor=red]{hyperref}
\usepackage[left=1.2in,right=1.2in,top=1.2in,bottom=1.2in]{geometry}
\usepackage{xcolor}
\usepackage{pdflscape}

\newcommand{\eg}{\textit{e.g.\ }}
\newcommand{\ie}{\textit{i.e.\ }}
\newcommand{\etal}{\textit{et al.\ }}

\begin{document}

\title{Hand Pose Estimation: A Survey}

\author{\textbf{Bardia Doosti}\\
\small{School of Informatics, Computing and Engineering}\\
\small{Indiana University Bloomington}\\
\small{Email: bdoosti@indiana.edu}}
\date{}

\maketitle

\begin{abstract}
The success of Deep Convolutional Neural Networks (CNNs) in recent years in almost all the Computer Vision tasks on one hand, and the popularity of low-cost consumer depth cameras on the other, has made Hand Pose Estimation a hot topic in computer vision field. In this report, we will first explain the hand pose estimation problem and will review major approaches solving this problem, especially the two different problems of using depth maps or RGB images. We will survey the most important papers in each field and will discuss the strengths and weaknesses of each. Finally, we will explain the biggest datasets in this field in detail and list 22 datasets with all their properties. To the best of our knowledge this is the most complete list of all the datasets in the hand pose estimation field.
\end{abstract}

\section{Introduction}
\begin{figure}[t]
\centering
    \includegraphics[scale=0.3]{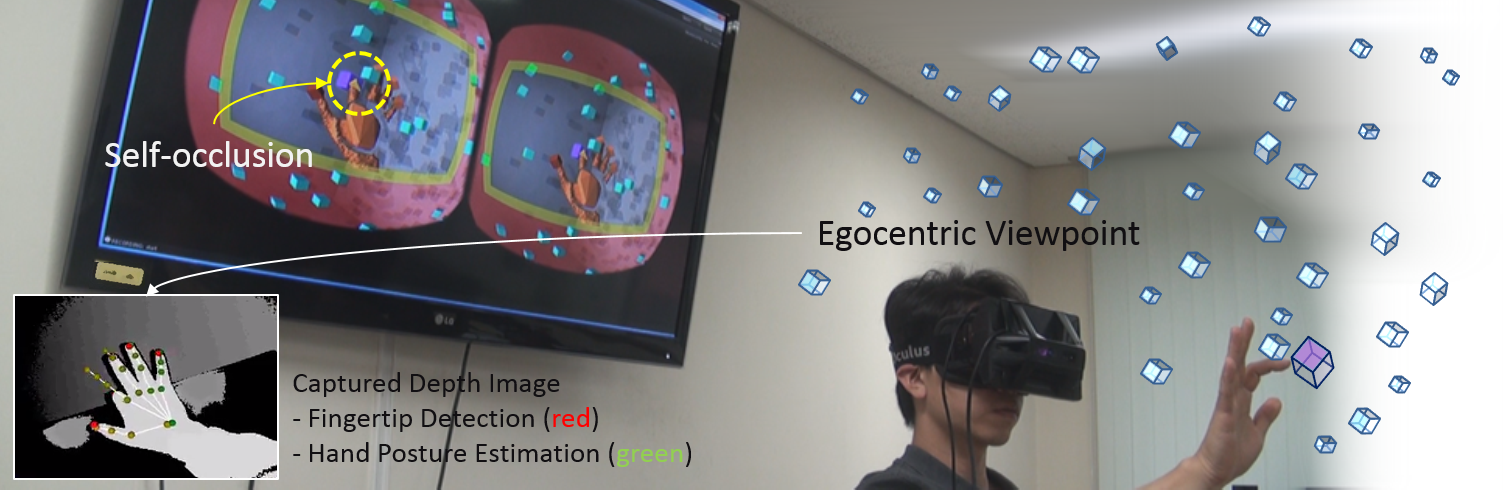}
    \caption{Application of hand pose estimation in egocentric viewpoint in AR/VR headset to control objects shown in the display. Originally used in~\cite{jang20153d}.}
    \label{fig:ar}
\end{figure}

Hand pose estimation is currently getting a lot of attention in the computer vision field. Since the invention of Deep Learning, researchers started to apply it in all computer vision fields and get a breakthrough result and hand pose estimation was not an exception. In addition the RGBD cameras which produce depth map have become cheap, which lowers the cost of making and using hand base systems. On the other hand, huge investment of big tech companies like Google, Microsoft and Facebook on Augmented Reality (AR), Virtual Reality (VR) and Mixed Reality (MR) technology as new interactive personal computers, has broadened the applications of this field. Consequently, a relatively new branch in Human Computer Interaction (HCI) has been introduced to study the systems controlled by understanding user's hands. In~\cite{lee2009multithreaded}, Lee \etal detected hands to render an object in AR environment on the hand which was proportional to hand size. Piumsomboon \etal~\cite{piumsomboon2013user} focused on {\it guessability} in 40 different tasks in AR environment with studying hand gestures. Jang \etal~\cite{jang20153d} build an AR/VR system in egocentric viewpoint which was completely controllable via user's hands. Figure~\ref{fig:ar} shows one of the applications of hand pose estimation used in egocentric viewpoint in AR/VR headset~\cite{jang20153d}.

Nevetheless the applications of hand pose estimation are not limited to AR/VR technologies. Sridhar \etal \cite{sridhar2015investigating} built a system working with a number of finger actions. Markussen \etal~\cite{markussen2014vulture} also proposed a mid-air keyboard to type on air. Yin \etal~\cite{yin2016iterative} used hand pose estimation to design a system that understands sign language. In a similar study, Chang \etal~\cite{chang2016spatio} used finger tip detection and tracking to read alphabet written by finger in the air. Outside HCI field, Shlizerman \etal and Rohrbach \etal applied body and hand pose estimation systems for predicting body movements of a piano player~\cite{Shlizerman_2018_CVPR} and to detect activities~\cite{rohrbach2012database}.

In recent years, the interest in systems controlled by fingers, made researchers more ambitious to the extent that they discarded 2.5D depth map images and tried to estimate hand pose by a single RGB image. This method is a harder task and needs a considerable larger data to train. Below, we will first explain the hand pose estimation problem and discuss its variations and next we will discuss different methods in solving this problem. At the end of the paper we will briefly investigate new datasets in this field and will see how the size of datasets have changed dramatically through time.

\section{Hand Pose Estimation Problem}
\begin{figure}[t]
\centering
\begin{tabular}{cc}
    \includegraphics[scale=0.4]{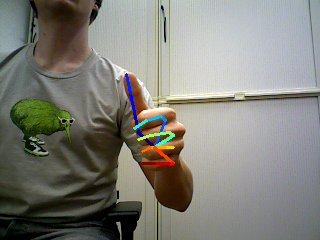} &
    \includegraphics[scale=0.22]{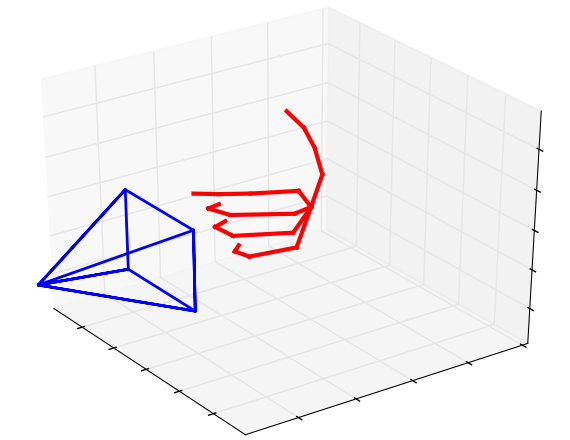} \\
    (a) & (b) 
\end{tabular}
\caption{(a) Image with 2D estimation of hand joints (b) 3D estimation of hand joints. Originally used in~\cite{zimmermann2017learning}.}
    \label{fig:estimate}
\end{figure}

Hand pose estimation is the process of modeling human hand as a set of some parts (\eg palm and fingers) and finding their positions in a hand image (2D estimation) or the simulation of hand parts positions in a 3D space
Although it is also used to estimate hand with the phalanges (like~\cite{wan2017hand} which is discussed in~\cite{Yuan_2018_CVPR} as \emph{strawberryfg} method), in almost all the recent papers hands are modeled as a number of joints and the task is equivalent to finding the position of these joints. We can then \emph{estimate} the real hand pose using those joints. Figure~\ref{fig:estimate} shows an image with its 2D and 3D estimation of hand pose using joints model connected with lines.

\begin{figure}[t]
\centering
\begin{tabular}{ccc}
    \includegraphics[scale=1.5]{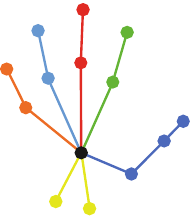} &
    \reflectbox{\includegraphics[scale=1.5]{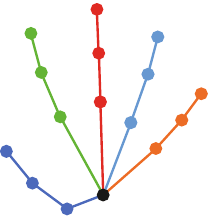}} &
    \includegraphics[scale=1.5]{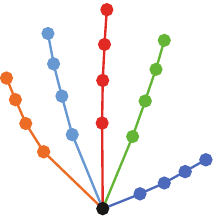} \\
    (a) 14 joints & (b) 16 joints & (c) 21 joints
\end{tabular}
\caption{Visualization of modeling hand with 14, 16 and 21 joints as used in NYU~\cite{tompson2014real}, ICVL~\cite{tang2014latent}, and MSRA~\cite{qian2014realtime} datasets. Originally used in~\cite{hu2018hand}.}
    \label{fig:joints}
\end{figure}

There is no global agreement on the number of joints used to model hands in different datasets. Those studies which want to compare their results for different datasets, have to change their model for each individual dataset. Albeit, 21 joints is now the most popular model and most of the datasets and pre-trained networks use this model. Figure~\ref{fig:joints} shows different number of joints in three popular hand datasets.

\section{Approaches}
\begin{figure}[t]
\centering
\begin{tabular}{ccc}
    \includegraphics[scale=0.115]{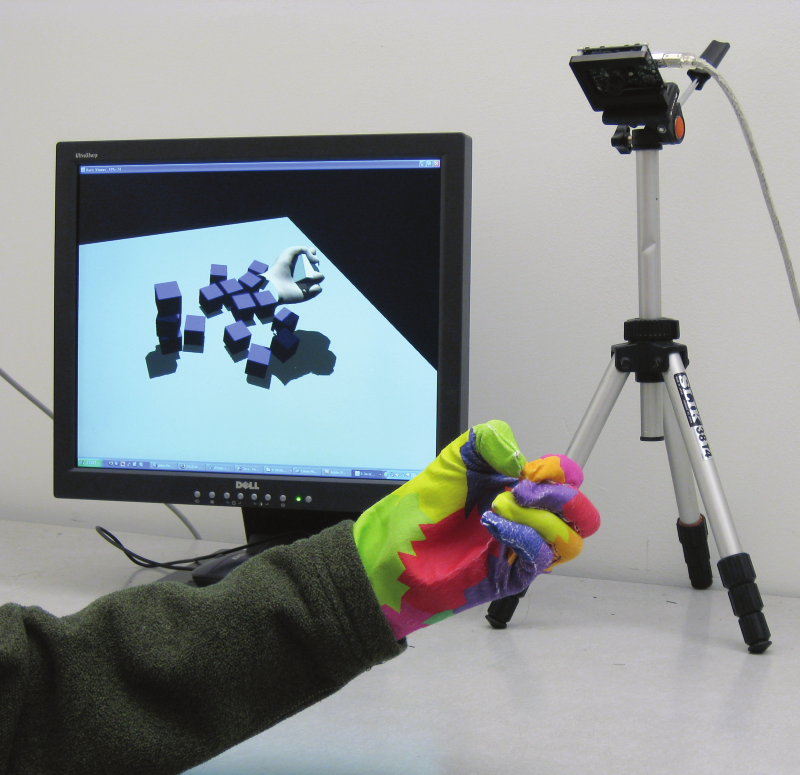} &
    \includegraphics[scale=0.3]{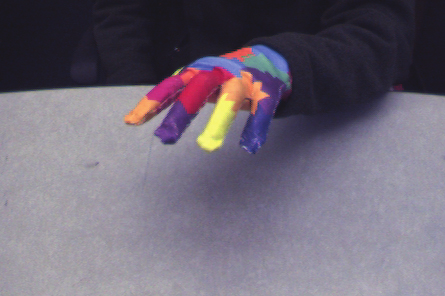} &
    \includegraphics[scale=0.2]{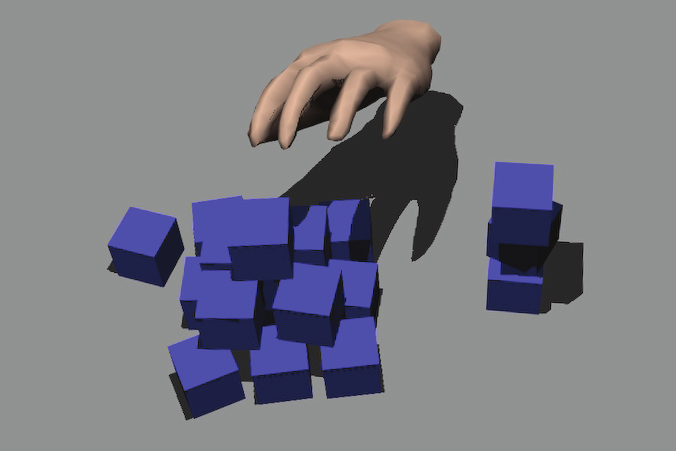}
\end{tabular}
\caption{Hand tracking using color glove. Originally used in~\cite{wang2009real}.}
    \label{fig:glove}
\end{figure}

Before deep learning revolution, people used to apply traditional machine learning and computer vision techniques for hand pose estimation. Wang \etal~\cite{wang2009real} used a color glove for user and then by using nearest neighbor they found the position of each color in the image and therefore the position of each specific part of the hand (Figure~\ref{fig:glove}). However, among all traditional approaches, random forest and its variations was the most popular one~\cite{tang2013real, tang2014latent, tang2015opening, sun2015cascaded, keskin2012hand, xu2013efficient}. At that time, this method was the most successful which made its way in the commercial products as well. Using depth camera, in Kinect, Microsoft applied random forest as a classifier for human body pose estimation~\cite{shotton2011real}. They first normalized depth map data using their neighbors value to be invariant to rotation. Then, they labeled each part of the body (similarly for hands) with a label and tried to classify each pixel (with its neighbor pixels) as one of these labels. Publishing a parallel algorithm to run decision tree and random forest quickly~\cite{sharp2008implementing}, they used random forest (which comprises a number of decision trees) to classify each point in the map.

In what follows, we will explain all the major approaches in solving hand pose estimation problem using deep learning. Although we can categorize these methodologies into estimating 2D or 3D skeleton, detection-based and regression-based algorithms, using 2D or 3D CNN, we divided them into methods using depth maps and methods using RGB image or both. We first define two different types of networks used in this problem which we will use in explaining algorithms.

\paragraph{Detection-based Methods \textit{vs.} Regression-based Methods}
In the detection-based method, the model produces a probability density map for each joint. So for example if a network uses 21-joints model for hands, for each image it will produce 21 different probability density maps as heatmaps. The exact location of each joint can be found by applying an \emph{argmax} function on corresponding heatmap. In contrast, regression-based method tries to directly estimate the position of each joint. That is, if it uses 21-joints model, it should have \(3 \times 21\) neurons in the last layer to predict \((x,y,z)\) coordinates of each joint. Due to the high non-linearity, training a regression-based network requires more data and training iterations. But since producing a 3D probability density function for each joint is a heavy task for a network, regression-based networks is used in 3D hand pose estimation tasks. Below we will discuss papers from both classes.

\begin{figure}[t]
\centering
    \includegraphics[scale=0.3]{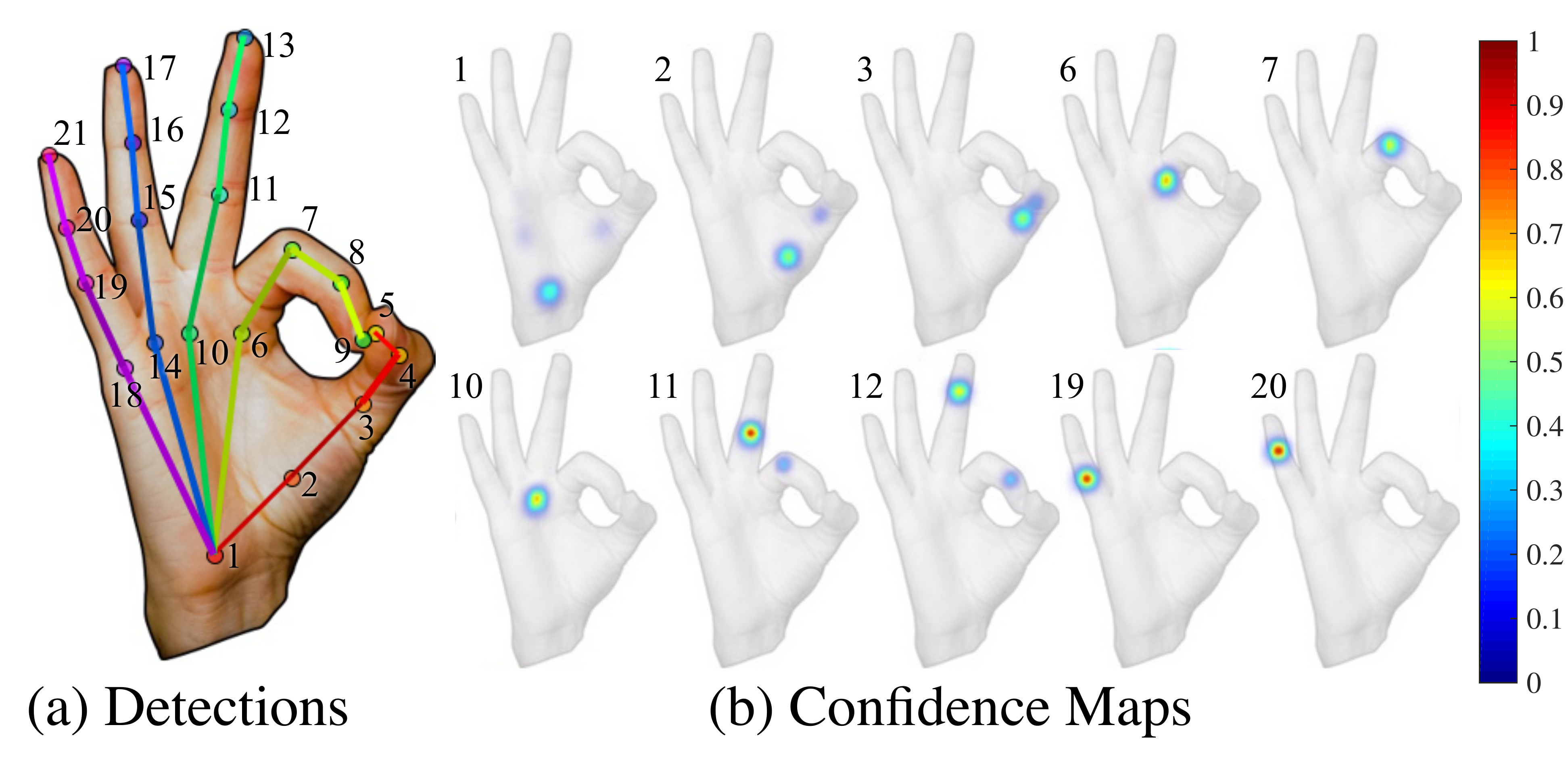}
\caption{The output of a detection-based algorithm. For each joint in the hand, one probability density function will be generated which is depicted as heatmaps. Originally used in~\cite{simon2017hand}.}
    \label{fig:detectionbased}
\end{figure}

\subsection{Depth-based Methods}
Traditionally, depth map image based methods were the main method in hand and body pose estimation. Sinha \etal~\cite{sinha2016deephand} used a regression-based method to find 21 joints in the hand, based on a depth map. They tried to find the location of joints in each finger independently. To this end, they trained a separate network for each finger to regress three joints on that finger. Note that although they used depth map to regress the coordinates, they also used RGB to isolate the hand and to remove all the other pixels included in the hand's cropped frame. Unlike the next papers that we will discuss, Sinha \etal did not use a separate deep network for hand segmentation, probably because of computation limitations. Instead they used RGB pixel color values to remove pixels which are not in the skin range.

\begin{figure}[t]
\centering
    \includegraphics[scale=0.9]{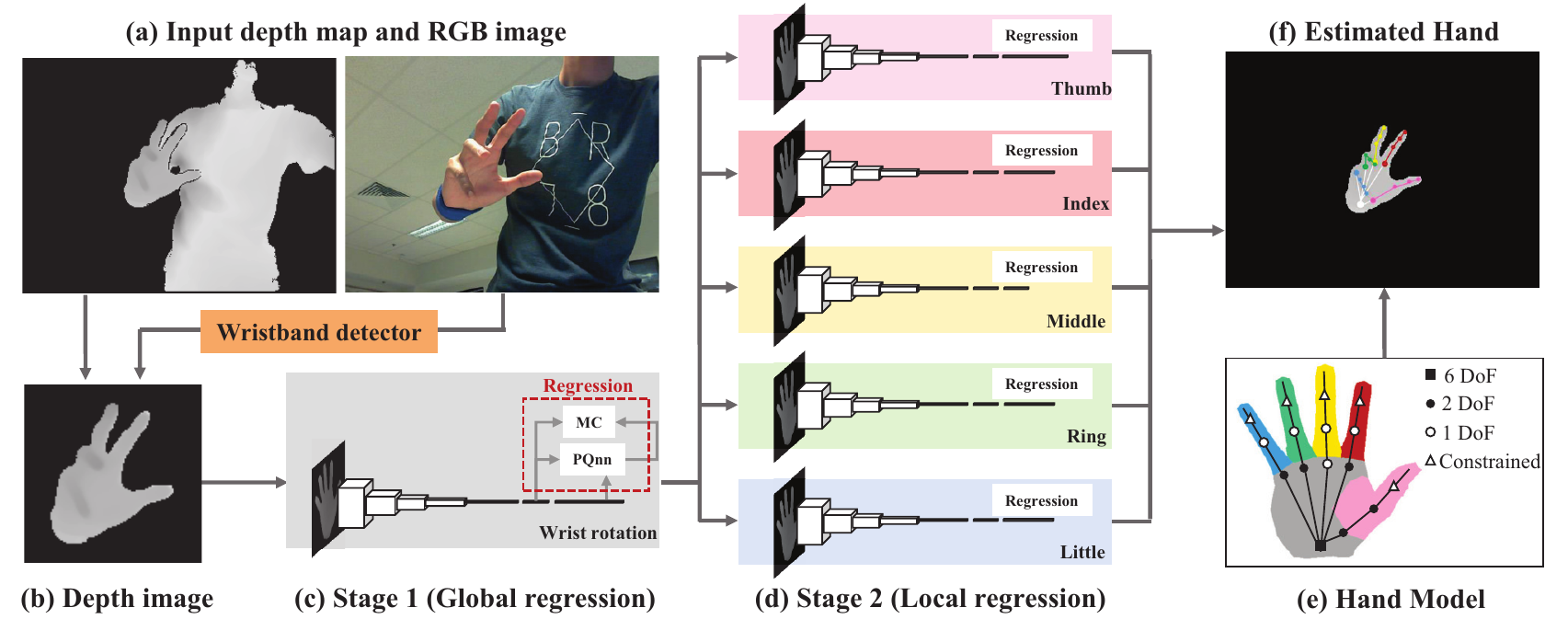}
\caption{Sinha \etal's multi network hand pose estimation. Originally used in~\cite{sinha2016deephand}.}
    \label{fig:multicnn}
\end{figure}

In~\cite{Baek_2018_CVPR} Baek \etal used Generative Adversarial Network (GAN)\cite{goodfellow2014generative} to estimate hand pose by making a one to one relation between depth disparity maps and 3D hand pose models. GAN is a specific CNN to generate new samples based on the previous learned samples. It consists of a discriminator and a generator network which are competing with each other to win the game. The discriminator network is a classifier trained to detect real and fake images. Generator is also a convolutional neural network which generates fake images based on a random initialization. These fake images should be good enough to deceive the discriminator as real ones. Conditioned GAN~\cite{pathak2016context} is a special GAN which gets a real image and it is conditioned to generate an image similar to that one; \ie the generator does not start from a random initialization.

\begin{figure}[t]
\centering
    \includegraphics[scale=0.88]{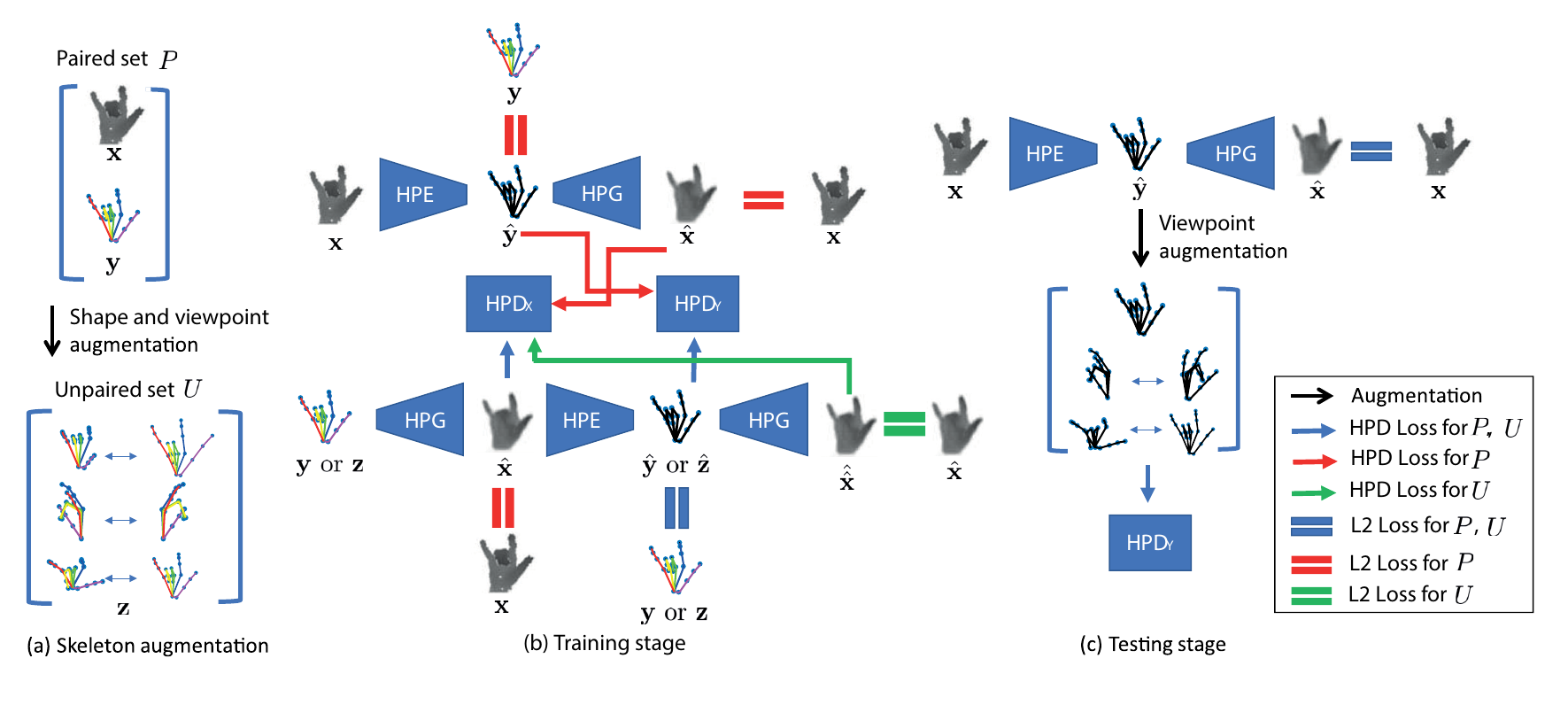}
\caption{Baek \etal's GAN based network architecture. In the diagram interaction with the paired set \(P\) and unpaired set \(U\) are represented by \textcolor{red}{Red} and \textcolor{green}{Green} respectively and the \textcolor{blue}{Blue} lines is for interaction with both \(U\) and \(P\). Originally used in~\cite{Baek_2018_CVPR}.}
    \label{fig:gan}
\end{figure}

Baek \etal used a CyclicGAN~\cite{zhu2017cyclic} which is a GAN for transferring one image from one domain to another. In this work one domain is the depth map of the hand and the other is the 3D representation of the hand joints. Baek \etal used a Hand Pose Generator (HPG) and a Hand Pose Discriminator (HPD) in their model. As you can guess from the above explanation, the HPG's job is to generate a hand, based on the 3D representation of the joints. In contrast, they used a Hand Pose Estimator (HPE) whose job is to generate the 3D hand pose, based on the input depth map. So in the training step HPG, HPD and HPE are optimized to reduce the error of HPE (which is the final goal of this algorithm) and to increase the consistency of the HPE-HPG combination \(f^E(f^G)): Y \rightarrow Y\) and the HPG-HPE combination \(f^G(f^E)): X \rightarrow X\). In the testing phase the algorithm refines the 3D model which is guided by the HPG to generate the best 3D model whose corresponding depth map is very similar to the input depth map. Figure~\ref{fig:gan} shows the schema of Baek \etal's algorithm.

Ge \etal~\cite{ge2016robust} created a new technique in the depth-based methods which was used in their next studies and by the other teams multiple times. The main idea of this paper (and their next paper~\cite{ge20173d}) is to \emph{estimate} 3D image from the 2.5D image and then estimate the hand pose from this new viewpoint. Although they did not mention how exactly this 3D model can be generated from depth-map, it can be understood from the cited papers, that they did not perform this part with machine learning algorithms.

As we know, depth maps only give us a surface of a hand, not a 3D shape. To estimate the 3D shape, they fix the camera in a 3D space and fix a surface as the farthest point in the camera's sight. So for every pixel in the depth map, proportional to the distance number, they should put voxels from that surface to the camera. With this method, the depth map produced from these voxels and the original depth map are the same. In the next step, they have to render these 3D volume from three perpendicular views; front, top and the side. To this end, they applied a Principle Component Analysis (PCA) on voxel's coordinates and picked the top three principal components and rendered 3D shape on those planes. They passed these depth maps to the network and got a probability for each joint in each \(xy\), \(xz\) and \(yz\) planes. In the next step which they call \emph{fusion} step, they mix the probabilities by multiplying them together.

\begin{figure}[t]
\centering
\begin{tabular}{ccc}
    \includegraphics[scale=0.55]{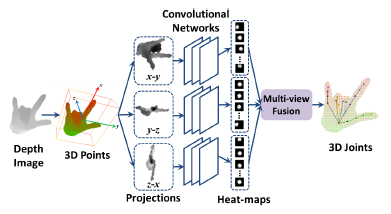} &
    \includegraphics[scale=0.4]{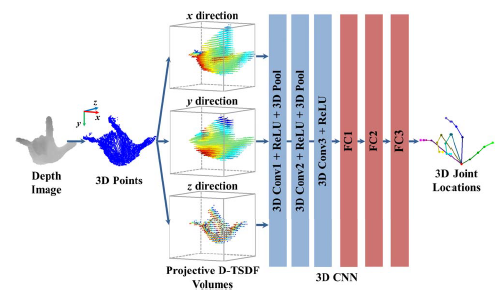} \\
    (a) & (b) 
\end{tabular}
\caption{Ge \etal's work in~\cite{ge2016robust} (a) and~\cite{ge20173d} (b). Originally used in~\cite{ge2016robust} and~\cite{ge20173d} respectively}
    \label{fig:multidepth}
\end{figure}

In their next paper~\cite{ge20173d}, Ge \etal performed the similar approach and used three different CNNs. But instead of 2D renders, they generated three 3D shapes with Truncated Signed Distance Function (TSDF). In TSDF shapes, signed distance of the voxel to the closest surface will be stored in each voxel. Also for the fusion step, instead of multiplying the probabilities, they concatenated the output vectors and used three fully connected layers to get the final result. They tested their method on MSRA~\cite{qian2014realtime} and NYU~\cite{tompson2014real} datasets and got the best results with a good margin comparing to other methods. Figure~\ref{fig:multidepth} shows schema of Ge \etal's~\cite{ge2016robust} and~\cite{ge20173d} papers models.

Inspired by Qi \etal's PointNet model in~\cite{qi2017pointnet++}, Ge \etal~\cite{Ge_2018_CVPR} applied PointNet on their 3D shape estimate (HandPointNet). First they sampled from points in the 3D shape to have a fixed and smaller number of points as the input. Also they applied their previous PCA analysis to rotate the 3D shape of the hand toward the principal components. Then they ran the hierarchical PointNet For 3D hand pose regression. In each step, this network downsampled the points. At last they used fully connected layers to regress the exact positions of hand joints in the space.

Ge \etal continued the same approach in~\cite{Ge_2018_ECCV} in which they changed the structure of their network. Therefore, instead of using multiple layers for downsampling the points, they used an architecture similar to encoder-decoder architecture. This structure first learns a global features and then using these global features it generates the desirable number of points used for estimating the position of joints.

Using a similar 2.5D depth map to 3D voxel-based hand shape convertor, in~\cite{Moon_2018_CVPR} Moon \etal designed a detection-based, voxel-to-voxel network (V2V) to directly estimate the position of each hand part based on the estimated 3D hand shape. Since the input and output are in 3D, they used 3D CNN which does all the convolution and deconvolution operations in 3D domain. The idea behind this paper is that depth maps taken from different angels of a single hand have the same 3D pose. To estimate hand pose via depth maps, it is needed to train the model to produce the same pose for different depth map inputs. On the other hand, a 3D point cloud has exactly one 3D hand pose and therefore their relation is one to one. So instead of having a huge dataset to cover all the shapes of a hand, we can train the model on the 3D point cloud of that hand and directly generate 3D pose via 3D encoder and decoders. 

\begin{figure}[t]
\centering
    \includegraphics[scale=0.65]{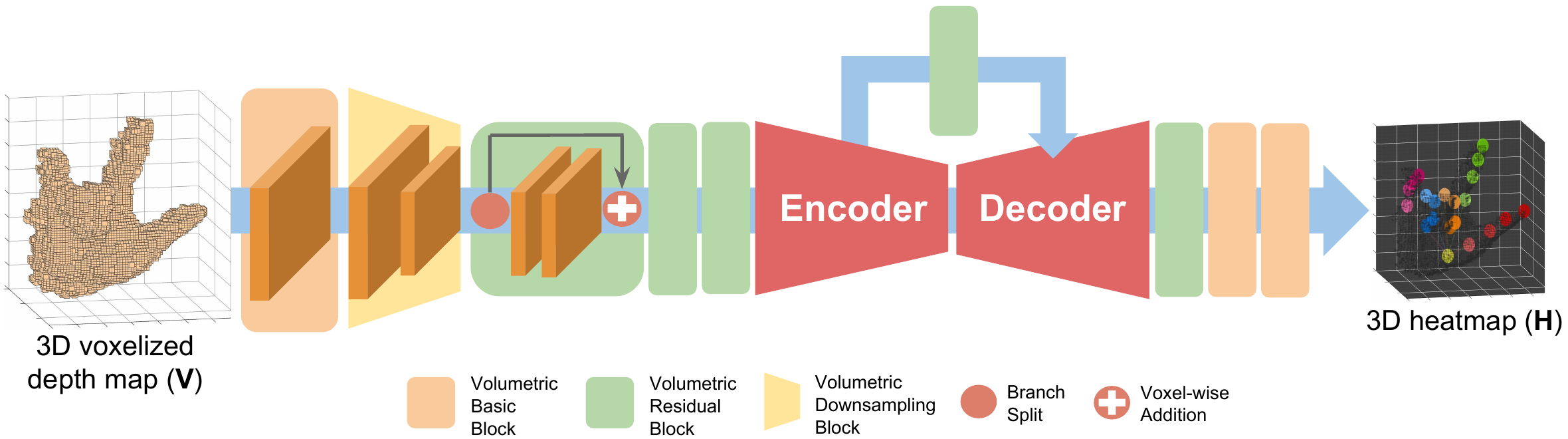}
\caption{Architecture of V2V-PoseNet network using 3D CNN as encoder and decoder. Originally used in~\cite{Moon_2018_CVPR}.}
    \label{fig:v2v}
\end{figure}

After the success of residual blocks of ResNet~\cite{he2016deep} in object classification, Moon \etal also used residual block with a deeper network. They applied their algorithm to almost all the famous depth-based hand datasets and compared their method with a vast majority of algorithms and got the best result with a high margin compared to the others. One of the interesting facts about this algorithm is that, it can be easily applied to body pose estimation problem as well. They tested their algorithm on ITOP dataset~\cite{haque2016towards} (both top-view and front-view) and reported their results. Figure~\ref{fig:v2v} shows the architecture of V2V-PoseNet.

\subsection{Image-based Methods}
In spite of the fact that using a simple RGB image as an input gives the model a very good generalization power to be used everywhere, reducing the dimension of the input from 2.5D to 2D will make the task drastically harder. The data needed to train a network using RGB images is much bigger than the data needed to train a similar network using depth maps. Below we will first discuss the general approach used in the RGB-based networks, then we will discuss how the top tier image-based algorithms solved the high cost of annotating data and making a bigger datasets. Therefore most of the influential studies in this section came up with their own dataset.

\begin{figure}[t]
\centering
    \includegraphics[scale=0.9]{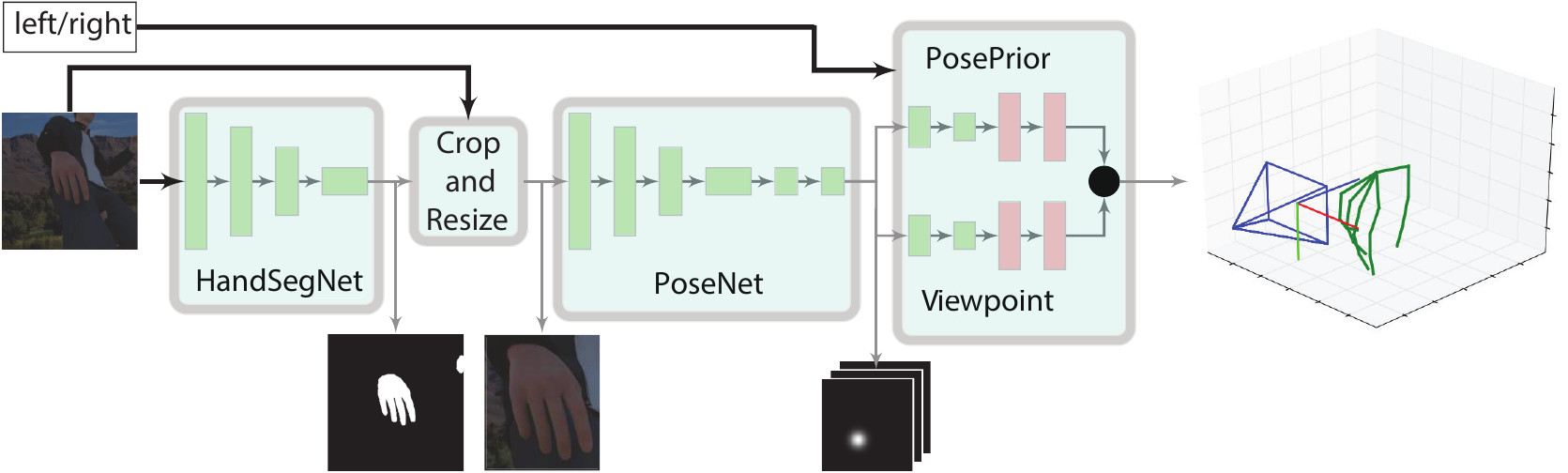}
\caption{Architecture of PoseNet network using detection-based network \emph{PoseNet} estimate 2D pose and regression-based network PosePrior to estimate the hand pose in 3D. Originally used in~\cite{zimmermann2017learning}.}
    \label{fig:posenet}
\end{figure}

Note that image-based methods need to first isolate the hand (crop and resize) and then pass the cropped image to the network to estimate the pose. To do this, most of the image-based networks use a variations of \emph{SegNet}~\cite{badrinarayanan2015segnet} which is designed to segment a general picture (\eg a street picture). Because of the binary classification (hand or background) and lack of variety in the input images, the hand segmentation is a relatively easier task than general segmentation problem. Therefore segmentation networks used in hand segmentation are more light weight than general SegNet and consequently faster to perform. In the following papers unless it is mentioned, one variation of SegNet is used for isolating the hands as a pre-processing step.

One of the important works in the RGB-based methods is Zimmermann \etal's paper and dataset~\cite{zimmermann2017learning}. In this paper they used four different deep learning streams to make a 3D estimation of hand joints using a single RGB image. They first used a CNN called \emph{HandSegNet} which is a light weight version of Wei \etal's~\cite{wei2016convolutional} human body detector trained on hand datasets.

As it is depicted in Figure~\ref{fig:posenet}, the output of \emph{HandSegNet} is a mask picture showing the hand pixels. Based on this image, the hand is cropped and resized and is passed to the \emph{PoseNet} network. PoseNet is a detection-based network which produces one probability density function (as a heatmap) for each hand joint. These predictions are in 2D and on the same coordinates of the input image. To get a 3D estimation, Zimmerman \etal used a network called \emph{PosePrior} to convert these 2D predictions to 3D hand estimation. This network is a regression-based network and predicts the coordinates of the joints followed by a normalization. In this step, they normalize the distances of the joints considering the farthest distance as 1 and dividing the other distances to that number. Finally, they find a 3D rotation matrix such that a certain rotated keypoints is aligned with y-axis of the canonical frame.

As mentioned earlier, since the RGB image contains less information than depth disparity maps, the RGB-based networka are harder to train and requires a larger dataset. The most important issue in image-based method is the occlusion case, that is when an object or the hand itself occludes some parts of the hand. The following two papers used two different innovative solutions to overcome this problem.

Inspired by Wei \etal's approach to estimate body pose, in~\cite{simon2017hand}, Simon \etal used a multicamera approach to estimate hand pose. They used Carnegie Mellon University's Panoptic Studio~\cite{Joo_2017_TPAMI} which contains more than 500 cameras (480 VGA and 30+ HD cameras) in a spherical space. They first trained a weak hand pose estimator using a synthesized dataset of hands. In the next step, they put a person in the center of the panoptic and applied the hand pose estimator on all the cameras recording video. The algorithm produces a (not very accurate) pose estimation for all of these views. It works in most of the views, but in the views in which the hand is occluded it does not work very well.

\begin{figure}[t]
\centering
    \includegraphics[scale=0.33]{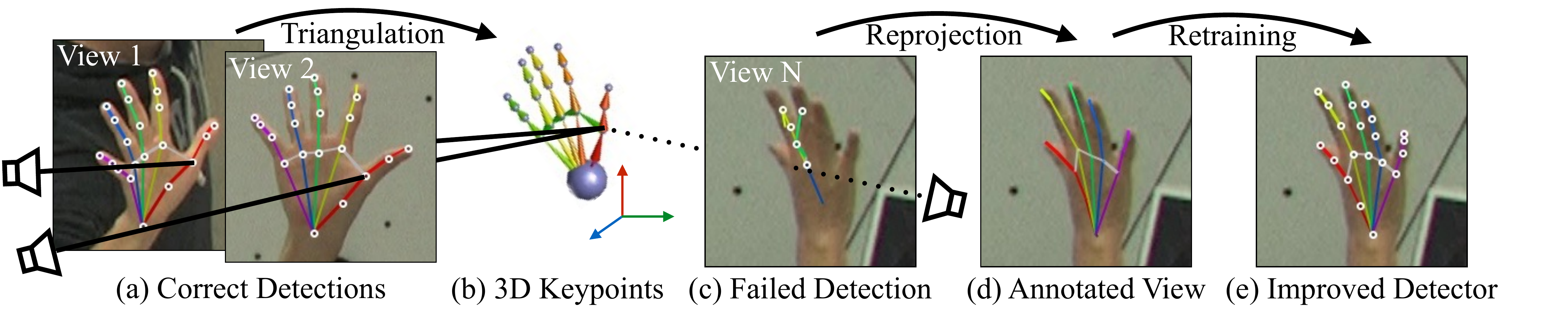}
\caption{the triangulation, projection and retraining steps in the Simon \etal's paper. Originally used in~\cite{simon2017hand}.}
    \label{fig:openpose}
\end{figure}

In the next part, which Simon \etal call triangulation step, they converted their 2D estimation to 3D estimation to evaluate their results, with knowing all the camera's intrinsic parameters and their relative physical position. To estimate the correct 3D pose, for each joint, they used RANSAC~\cite{fischler1981random} algorithm to randomly select 2D views and convert them to 3D view. Then they keep the model with which most of the views agree. Finally, in a reverse operation, they projected the 3D view to the pictures and annotate that frame. In fact, instead of annotating data manually, they used multiple views to annotate their data. Then with this annotated dataset which is from multiple views, they trained the network again and made it more accurate. They repeated this process of annotation and training the model for three times and therefore they ended up with a very good and accurate model and dataset of annotated hand poses. For the feature extraction, they used a pre-trained VGG-19 network~\cite{simonyan2014very} (up to \(\text{conv}4\_4\)) which produces a 128-channel feature. Figure~\ref{fig:openpose} shows the triangulation, projection and retraining steps in the Simon \etal's~\cite{simon2017hand} paper.

To overcome the occlusion problem and especially the size of the dataset and the high cost of annotating hands frame by frame, Mueller \etal~\cite{Mueller_2018_CVPR} used a synthesized dataset which is annotated automatically. They used kinematic sensors which has multiple electromagnetic sensors (usually 6 6D sensors) connected to a hand. These sensors are connected to a receiver and a transmitter which generates the 3D hand pose automatically.

\begin{figure}[t]
\centering
    \includegraphics[scale=0.45]{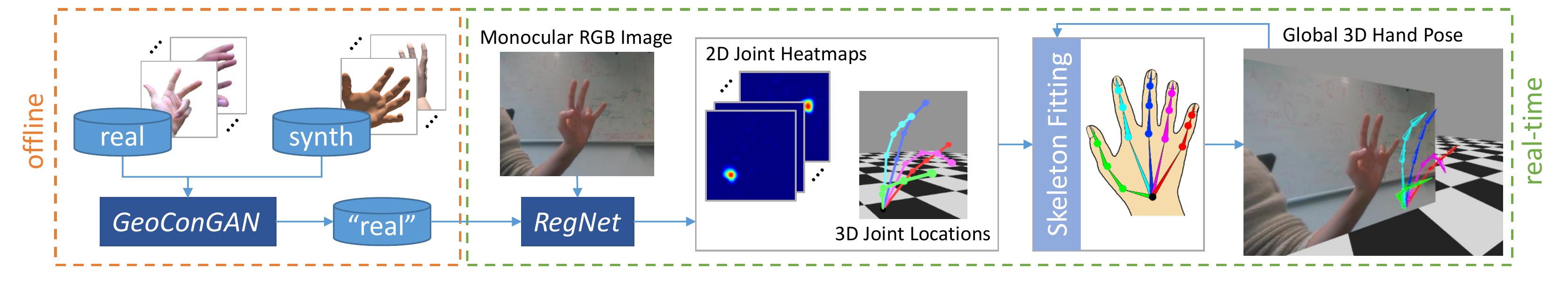}
\caption{different steps of dataset production and hand pose estimation of Mueller \etal's paper. Originally used in~\cite{Mueller_2018_CVPR}.}
    \label{fig:ganerated}
\end{figure}

Although using synthesized dataset is easy to generate and annotate, they lack generalization power. As the image generated by this devices are computer generated, it will not work very well on real-world hand images. To overcome this issue they used a conditioned GAN~\cite{pathak2016context} called \emph{GeoConGAN} to transfer the computer generated images to real images. Also, to reach a better one-to-one relation between real and computer generated images, they applied a CyclicGAN~\cite{zhu2017cyclic} which has two parts of Real to Synthesized GAN (called \emph{real2synth}) and Synthesized to Real GAN (called \emph{synth2real}). Each of this GANs has its own generator and discriminator. Mueller \etal controlled the process with two losses; first converting synthesized image to real and calculating synth2real loss and again converting the result to synthesized image and calculating real2synth loss. They also randomly put some backgrounds behind the hands to make the images more realistic. Moreover, to create occlusion on the hands, they artificially put some objects in front of the hands to have some occluded frames in the dataset as well. They used ResNet~\cite{he2016deep} architecture for their feature extraction network to take advantage of residual blocks. Figure~\ref{fig:ganerated} shows the different steps of dataset production and hand pose estimation of Mueller \etal's paper~\cite{Mueller_2018_CVPR}.

Spurr \etal~\cite{Spurr_2018_CVPR} also used this cyclic concept for making a one-to-one relation between RGB image to 3D hand joints pose. They used GAN and 
Variational Autoencoder (VAE) to transfer the images to a latent space and then transfer it to the other domain. With that, they tried to map every RGB hand image to a 3D pose and use this map in the hand pose estimation task.

So far we only discussed algorithms which have using depth disparity maps or single RGB image. But there are studies in which RGBD images have been used (\ie using both RGB image and its corresponding depth disparity map). Moreover some researchers used RGBD images during the training and used RGB while testing.

Dibra \etal in~\cite{dibra2018monocular} designed a network which uses both RGB images and depth maps to estimate the hand pose. They used a specific network called \emph{SynthNet} to estimate the hand pose which will be explained shortly. Next, with the generated 3D shape (3D shape of the whole hand not just the skeleton) they generate a depth map. They did the same process for depth map input as well. With the method explained above they generated a 3D model from the 2.5D depth map and then they calculated the loss of the algorithm from difference of two depth maps generated from these 3D models.

\begin{figure}[t]
\centering
    \includegraphics[scale=0.9]{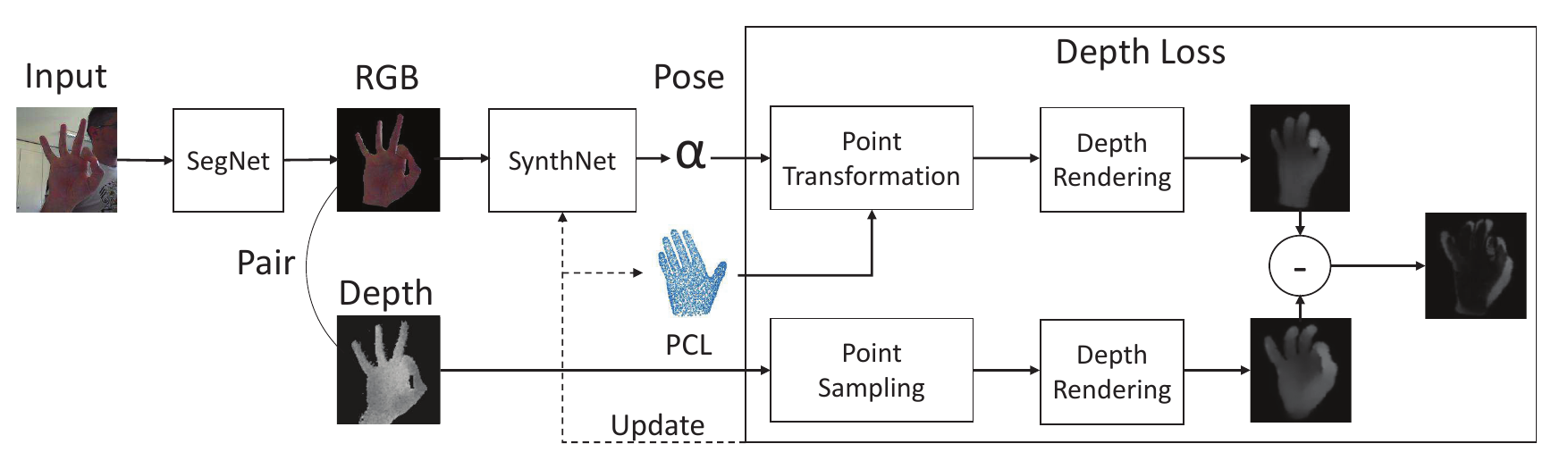}
\caption{Different stages in Dibra \etal's 3D hand pose estimation algorithm. Originally used in~\cite{dibra2018monocular}.}
    \label{fig:dibra}
\end{figure}

Unlike almost all the papers discussed so far, Dibra \etal did not use a joint model. In this paper they came up with a new method for hand pose estimation which unlike the other methods is not data-driven but uses information about human hand anatomy. So in their model they produce a hand shape which is physically possible (given all the possible cases of hand pose in prior). Figure~\ref{fig:dibra} shows different stages in Dibra \etal's 3D hand pose estimation algorithm.

\begin{figure}[t]
\centering
    \includegraphics[scale=0.9]{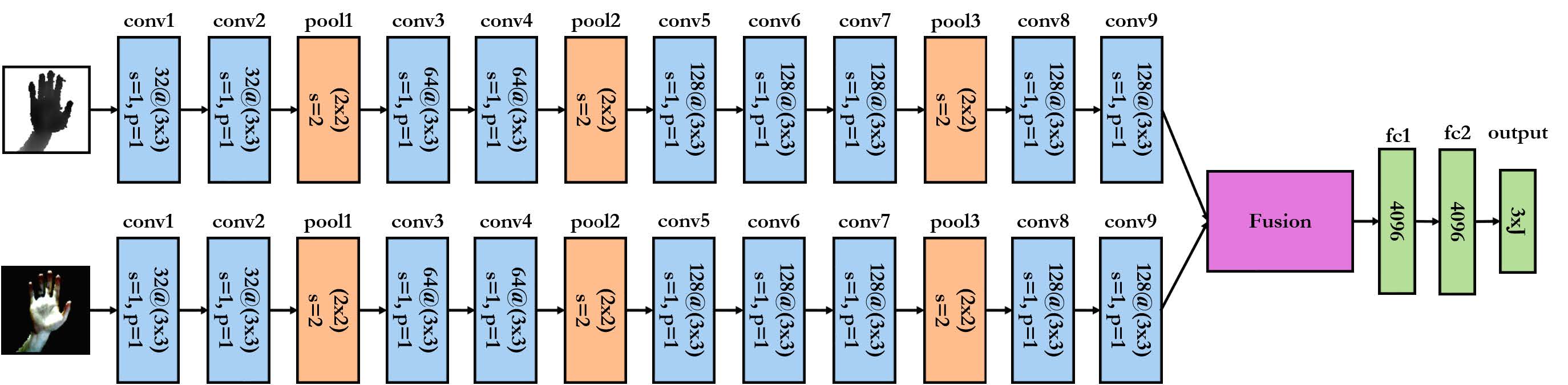}
\caption{The two different streams in the architecture of \emph{FuseNet}. Originally used in~\cite{kazakos2018fusion}.}
    \label{fig:both}
\end{figure}

Also there are models which used both depth disparity maps and RGB images. To make estimation more accurate, Kazakos \etal in~\cite{kazakos2018fusion} used two different deep learning streams for RGB and depth disparity maps called \emph{FuseNet}. They used two identical convolutional neural networks for feature extraction of RGB and depth map images. The final prediction is generated from two fully connected layer which regresses \((x,y,z)\) coordinates of each joint. Despite using two different streams, they results was not as good as the other approaches using one of these inputs. Figure~\ref{fig:both} shows the two different streams in the architecture of \emph{FuseNet}.

\section{Datasets}
In this section, we explain some of the most important datasets used in hand pose estimation and discuss their properties in detail. Also you can find the list of 22 hand datasets in Table~\ref{tab:datasets} which, to the best of our knowledge, is the most complete list of all the datasets in the hand pose estimation field.

\subsection{ICVL Hand Dataset}
The Imperial College Vision Lab (ICVL) dataset~\cite{tang2014latent} is one of the oldest datasets in hand pose estimation field. It contains 180K annotated depth frames. They used 10 subjects to take 26 different poses. 16-joints model was used in annotating this dataset.

\subsection{NYU Hand Dataset}
New York University (NYU) dataset~\cite{tompson2014real} contains 72,757 frames from a single actor in the train set and 8,252 frames from two different actors in test set. It is an RGBD dataset captured from side view (3rd person view). 36-joints model was used model to annotate this dataset.

\subsection{HandNet Dataset}
HandNet dataset~\cite{wetzler2015rule} is one of the biggest depth datasets. It is generated using kinematic sensors with 10 different subjects, half male and half female to have different hand sizes in the dataset. It contains 202K frames in the trainning set and 10K frames in the test set. 6-joints model was used to annotate data.

\subsection{CMU Panoptic Hand Dataset}
Carnegie Mellon University (CMU) Panoptic~\cite{simon2017hand} is RGB images which are recorded and annotated in the CMU's Panoptic studio. It contains both the synthesized and real images with 14,817 frames in 3rd person view which are annotated by 21-joints model in 2D.

\subsection{BigHand 2.2M Benchmark Hand Dataset}
The BigHand 2.2M Benchmark~\cite{yuan2017bighand2} is the biggest hand dataset so far. As we can see from its name it has, 2.2M annotated depth frames which are generated from 10 different subjects using kinematic six 6D electromagnetic sensors. Like all the recent datasets, it uses 21-joints model annotated in 3D. As the whole dataset was annotated with kinematic sensors, no object was held in the hands.

\subsection{First-Person Hand Action dataset}
The First-Person Hand Action dataset (FHAD)~\cite{garcia2018first} is a new dataset introduced by The Imperial College. It contains 105,459 frames of egocentric view of 6 subjects doing 45 different types of activities in the kitchen, in the office or social activities. This dataset was also annotated in 3D, using 21-joints model.

\subsection{GANerated Hand Dataset}
GANerated is a new big dataset for the RGB-based dataset which has interaction with objects that can be helpful in estimating the hand pose under occlusion. It contains 330K frames synthesized hand shapes annotated in 3D using 21 joints model. In this dataset kinematic electromagnetic sensors was used to capture the hand pose. Also a CycleGAN was utilized to convert these computer generated images to look like real images. Different backgrounds was randomly put behind the hands to make them more similar to real photos. Also artificial objects was put on the hand to produce a hand occlusion.

The complete list of hand datasets with all their major properties is listed in Table~\ref{tab:datasets}.

\section{Conclusion}
In this report we defined the hand pose estimation problem and explained the major methods of solving this problem in detail. We also reviewed some of the recent applications of this field. Since every data driven method needs sufficient data in the first place, we talked about major datasets and listed all the datasets in this field with their most important properties. We showed how this field have grown in just a few years, from completely controlled situations with color gloves to 3D hand pose estimation using a single RGB image. Although the papers discussed here show good results on these dataset they do not get satisfactory results in the real world problems. Most importantly, the result of most of these systems is worst than a simple nearest-neighbor baseline~\cite{Yuan_2018_CVPR}. However, because of the interests of big technology companies in this field, perhaps in the near future we see much bigger and more generalized datasets and therefore very well performing models even on a single RGB image. If we reach this technology, using an AR/VR device as our new PC, typing on the air and control objects in the display with our fingers will not be out of reach to .

\begin{landscape}
\begin{table}
\centering
\label{tab:datasets}
\caption{List of hand datasets with their properties}
\renewcommand{\arraystretch}{1.3}
\begin{tabular}{|c|c|c|c|c|c|c|c|c|c|}
\hline
Dataset &Year & Synth./Real & RGB/D & Objects & \#Joints & View & Ann. & \#Subjects & \#Frames (train/test) \\\hline\hline
Occluded Hands~\cite{myanganbayar2018partially} & 2018 & Real & RGB & Yes & 21 & 3rd & 2D & 24 & 11,840 \\\hline
GANerated~\cite{Mueller_2018_CVPR} & 2018 & Synth. & RGB & Yes & 21 & Ego & 2D+3D & - & 330,000\\\hline
FHAD~\cite{garcia2018first} & 2018 & Real & RGB+D & Yes & 21 & Ego & 3D & 6 & 105,459\\\hline
BigHand2.2M~\cite{yuan2017bighand2} & 2017 & Real & D & No & 21 & 3rd & 3D & 10 & 2.2M\\\hline
EgoDexter~\cite{mueller2017real} & 2017 & Real & RGB+D & Yes & 5 & Ego & 3D & 4 & 1,485\\\hline
SynthHands~\cite{mueller2017real} & 2017 & Synth. & RGB+D & Both & 21 & Ego & 3D & 2 & 63,530\\\hline
RHD~\cite{zimmermann2017learning} & 2017 & Synth. & RGB+D & No & 21 & 3rd & 3D & 20 & 41k/2.7k\\\hline
STB~\cite{zhang2017hand} & 2017 & Real & RGB+D & No & 21 & 3rd & 3D & - & 18,000\\\hline
CMU Panoptic~\cite{simon2017hand} & 2017 & Both & RGB & No & 21 & 3rd & 2D & - & 14,817\\\hline
Graz16~\cite{oberweger2016efficiently} & 2016 & Real & RGB+D & Yes & 21 & Ego & 2D+3D & 6 & 2,166\\\hline
Dexter+Object~\cite{sridhar2016real} & 2016 & Real & RGB+D & Yes & 5 & 3rd & 3D & 2 & 3,014\\\hline
EgoHands~\cite{Bambach_2015_ICCV} & 2015 & Real & RGB & Yes & - & Ego & - & 48 & 15,053\\\hline
MSRA15~\cite{sun2015cascaded} & 2015 & Real & D & No & 21 & 3rd & 3D & 9 & 76,375\\\hline
MSRC~\cite{sharp2015accurate} & 2015 & Synth. & D & No & - & - & 3D & - & -\\\hline
HandNet~\cite{wetzler2015rule} & 2015 & Real & D & No & 6 & 3rd & 3D & 10 & 202k/10k\\\hline
NYU~\cite{tompson2014real} & 2014 & Real & D & No & 36 & 3rd & 3D & 2 & 72k/8k\\\hline
ICVL~\cite{tang2014latent} & 2014 & Real & D & No & 16 & 3rd & 3D & 10 & 331k/1.5k\\\hline
UCI-EGO~\cite{rogez20143d} & 2014 & Real & RGB+D & No & 26 & Ego & 3D & 2 & 400\\\hline
Hands in Action~\cite{tzionas2016capturing} & 2014 & Real & D & Yes & - & 3rd & 3D & - & -\\\hline
MSRA14~\cite{qian2014realtime} & 2014 & Real & D & No & 21 & 3rd & 3D & 6 & 2,400\\\hline
Dexter1~\cite{sridhar2013interactive} & 2013 & Real & RGB+D & No & 6 & 3rd & 3D & 1 & 2137\\\hline
ASTAR~\cite{xu2013efficient} & 2013 & Real & D & No & 20 & 3rd & 3D & 30 & -\\\hline
\end{tabular}
\end{table}
\end{landscape}

\bibliography{report}{}
\bibliographystyle{plain}

\end{document}